\documentclass[lettersize,journal]{IEEEtran}
\usepackage{amsmath,amssymb,amsfonts}
\usepackage{bm}
\usepackage{graphicx}
\usepackage{cite}
\usepackage{algorithmic}
\usepackage{algorithm}
\usepackage{array}
\usepackage{textcomp}
\usepackage{verbatim}
\usepackage{url}
\usepackage{hyperref}
\usepackage[table]{xcolor}
\usepackage{booktabs}
\usepackage{multirow}
\usepackage{subcaption}
\usepackage{siunitx}
\usepackage{pifont}
\usepackage{tikz}
\usepackage{stfloats}
\newcommand{\dq}{\hat{\bm{q}}}

\newcommand{\se}{\mathfrak{se}(3)}
\newcommand{\so}{\mathfrak{so}(3)}
\newcommand{\SE}{SE(3)}
\newcommand{\SO}{SO(3)}

\newcommand{\R}{\mathbb{R}}
\newcommand{\Hd}{\mathcal{H}}
\newcommand{\Hdp}{\mathcal{H}_p}
\newcommand{\Sd}{\mathcal{S}}

\DeclareMathOperator{\Log}{Log}

\newcommand{\vect}[1]{\bm{#1}}

\newcommand{\dt}{\Delta t}

\begin{document}

\newif\ifanon
\anonfalse

\newif\ifmarkrevr
\markrevrfalse
\newcommand{\revr}[1]{\ifmarkrevr{\color{blue}#1}\else{#1}\fi}

\setlength{\floatsep}{4pt plus 1pt minus 1pt}
\setlength{\textfloatsep}{5pt plus 1pt minus 1pt}
\setlength{\dblfloatsep}{4pt plus 1pt minus 1pt}
\setlength{\dbltextfloatsep}{5pt plus 1pt minus 1pt}
\renewcommand{\topfraction}{0.95}
\renewcommand{\dbltopfraction}{0.95}
\renewcommand{\floatpagefraction}{0.9}
\renewcommand{\dblfloatpagefraction}{0.9}
\renewcommand{\textfraction}{0.05}
\setcounter{totalnumber}{6}\setcounter{topnumber}{4}\setcounter{dbltopnumber}{3}

\newcommand{\DQMPCC}{\textbf{DQ-MPCC}}
\title{\LARGE \bf
DQ-MPCC: Dual-Quaternion MPCC for Quadrotor Racing
}
\ifanon
\author{Anonymous Author(s)%
\thanks{Submitted to IEEE Robotics and Automation Letters. Author names,
        affiliations, funding, and the code repository are omitted for
        double-anonymous review.}}
\markboth{}{}
\else
\author{Bryan~S.~Guevara$^{\dagger}$,
        Luis~F.~Recalde$^{\dagger}$, Guanrui Li,
        and~Tiago~Nascimento
\thanks{Submitted to IEEE Robotics and Automation Letters.}
\thanks{$^{\dagger}$These authors contributed equally to this work.}
\thanks{Bryan~S.~Guevara and Tiago~Nascimento are with the
        Laboratory of Systems and Robotics Engineering (LASER),
        Center of Informatics (CI), Federal University of
        Paraíba (UFPB), João Pessoa, PB 58051-900, Brazil.
        {\tt\small \{bguevara, tiago\}@laser.ufpb.br}}
\thanks{Luis~F.~Recalde and Guanrui Li are with the Department of Robotics
        Engineering, Worcester Polytechnic Institute (WPI),
        Worcester, MA 01609, USA.
        {\tt\small {lfrecalde, gli7}@wpi.edu}}
\thanks{This work has been supported by Petrobras}}
\markboth{}{}
\fi


\maketitle

\begin{abstract}
Quadrotor racing demands \revr{aggressive attitude and progress control} while passing through every gate, and \revr{conventional
quadrotor MPCC formulations state the prediction model in inertial coordinates and the attitude error in the body frame.}
We present a Dual-Quaternion Model Predictive Contouring
Control (\DQMPCC{}) for quadrotor racing in which the pose is a unit dual
quaternion and the contouring errors are projected onto the tangent space of
the dual quaternion manifold, expressed in the desired body frame:
\revr{the same rigid-body dynamics as the conventional model, in unified
pose-twist coordinates in the body frame.} We compare \DQMPCC{} against a
baseline MPCC through Monte Carlo software-in-the-loop simulations and
real-world racing on an eight-gate circuit of $11\times4.5\times3.65$\,m.
\revr{With the same gains in simulation and hardware, \DQMPCC{} keeps every crossing of its
completed flights within the prescribed geometric margin, whereas the baseline exceeds it, its median worst-gate offset growing by $71.5\%$ sim-to-real against a $10.1\%$ decrease for \DQMPCC{}.} \revr{Among the configurations that keep every simulated gate crossing within the geometric margin,} \DQMPCC{} also reduces the minimum lap time by $6.7\%$, and by $10.5\%$ in the real-world flights, while running onboard at $100$\,Hz.
\end{abstract}

\ifanon\else
\fi

\section{Introduction}
\label{sec:intro}

\ifanon
\revr{Minimum-time}
\else
\IEEEPARstart{M}{\revr{inimum-time}}
\fi
flight through a gate sequence is the standard autonomous racing benchmark~\cite{foehn2021time, kaufmann2023champion, song2023reaching, hanover2024survey}.
Controllers operate \revr{under strong translational and rotational coupling}, while still
passing through every gate~\cite{penicka2022minsnap}.
Nonlinear MPC now runs onboard in real time with performance
competitive with learned racing
policies~\cite{kaufmann2023champion,song2023reaching,hanover2021l1nmpc}.

Model Predictive Contouring Control (MPCC)~\cite{liniger2015optimization,lam2010mpcc} has become a standard formulation for minimum-time racing because it avoids prescribing a fixed time allocation along the reference path. MPCC optimizes the path progression online, advancing as fast as the constraints allow~\cite{romero2022model,cmpcc2021,mpccpp2024}. Existing quadrotor MPCC formulations typically express the translational contouring error in inertial coordinates and the attitude error in a separate rotational representation~\cite{romero2022model,cmpcc2021,mpccpp2024}, MPCC++~\cite{mpccpp2024} included, which adds a terminal safety set, learned residual dynamics, and Bayesian tuning\revr{; position and attitude remain physically coupled through the dynamics.}

\revr{Under high-tilt maneuvers}, reducing an inertial position error may require rapid attitude changes, so the position and attitude objectives are coupled where gate margins are tightest. \revr{Both coordinate choices describe the same physical system; we compare the two formulations under one tuning and deployment protocol.}

\begin{figure}[!t]
    \centering
    \includegraphics[width=\columnwidth]{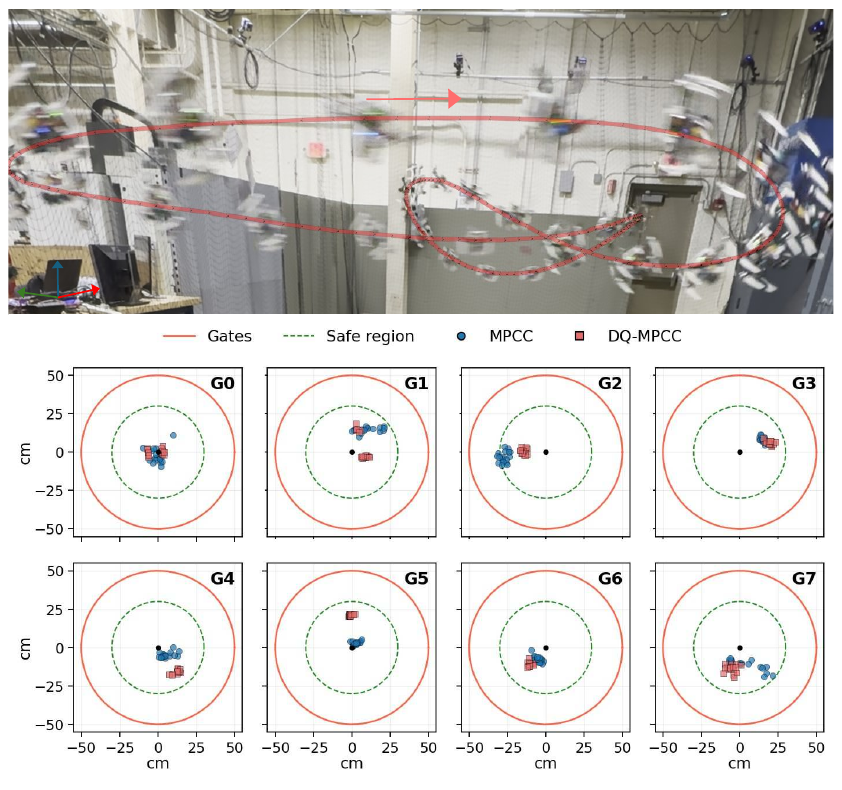}
    \caption{Real-world experiment of the proposed
    \DQMPCC{} through the $8$-gate circuit (top). The bottom
    plots show the gate crossings of all racing laps, with red
    markers for the \DQMPCC{} and blue markers for the
    MPCC baseline.}
    \label{fig:portada}
\end{figure}

Unit dual quaternions \revr{provide a unified rigid-body pose representation whose exact logarithmic map represents translational and rotational errors jointly} in a tangent space isomorphic to $\se$, the Lie algebra of the pose group $\SE$~\cite{selig2005,adorno2017fundamentals}. Geometric controllers built on this representation enable singularity-free tracking of coupled rigid-body motion~\cite{wang2013dual,abaunza2017}. \revr{A related line formulates MPC directly on matrix Lie
groups~\cite{kalabic2017mpcso3,teng2022errorstate}; we bring this
on-manifold perspective to contouring control.}
Recent work on DQ-NMPC~\cite{recalde2025dqnmpc} has further demonstrated the effectiveness of dual-quaternion formulations for aggressive quadrotor trajectory tracking. Despite these advances, dual-quaternion formulations have not, to our knowledge, been investigated for \revr{minimum-time} quadrotor racing. \revr{Relative to the time-indexed tracking objective of DQ-NMPC~\cite{recalde2025dqnmpc}, DQ-MPCC makes arc-length progress a decision variable and decomposes the translational pose-error coordinate along the path tangent into separately weighted lag and contouring components, on which the soft lag and variable-radius tube constraints act.}


We present \DQMPCC{}, a dual-quaternion MPCC formulation in which contouring error terms are obtained by projecting the pose error onto the tangent space of the unit dual-quaternion manifold and expressing it in the desired body frame. Our method is compatible with standard flight controllers and runs onboard in real time at $100$\,Hz \revr{with an external state estimate}. We evaluate \DQMPCC{} against an MPCC baseline with the conventional \revr{inertial position-error formulation} on an $8$-gate circuit in both Software-in-the-Loop (SiL) simulation, in which the onboard flight computer runs the unmodified control software while the vehicle, its state estimate, and the actuator interface are simulated in MuJoCo, and real-world (RW) experiments.

As shown in Table~\ref{tab:comparison}, \revr{the compared} MPCC and dual-quaternion methods do not formulate \revr{minimum-time} quadrotor racing directly on $\SE$. To close this gap, this paper makes the following three contributions\revr{, building on the dual-quaternion modeling of DQ-NMPC~\cite{recalde2025dqnmpc}}:
\begin{itemize}
    \item We introduce \DQMPCC{}, a racing controller formulated
    directly on the unit dual-quaternion manifold, whose prediction model
    is the dual-quaternion dynamics in the body frame. \revr{The decomposition into lag and contouring errors is defined in the tangent space of the pose error, with
    pose-error, twist, and prediction quantities expressed in body-fixed coordinates; the deployed controller evaluates it with the identity approximation.}
    \item \revr{We demonstrate through Monte Carlo SiL simulations that, among
    the configurations that keep every gate crossing within the geometric margin, \DQMPCC{}
    reduces the lap time by $6.7\%$ compared with a conventional MPCC baseline using an inertial position-error formulation,} while maintaining real-time execution.
    \item We evaluate \DQMPCC{} in real-world experiments on an eight-gate circuit.
    \revr{Without sim-to-real retuning, \DQMPCC{} keeps every gate-crossing offset of its completed flights within the prescribed geometric margin, whereas the baseline exceeds that margin at its worst gate;} the minimum lap time is reduced by $10.5\%$ compared with the baseline MPCC.
\end{itemize}

\section{Dual Quaternion Modeling of the Quadrotor}
\label{sec:model}

\subsection{Dual Quaternion Algebra}
\label{sec:dq_algebra}

\revr{We summarize only what the formulation uses; see~\cite{selig2005,adorno2017fundamentals} and DQ-NMPC~\cite{recalde2025dqnmpc} for
complete treatments.} Quaternions
$\bm{q}=q_w+x\mathbf{i}+y\mathbf{j}+z\mathbf{k}\in\mathbb{H}$
\revr{have} \revr{scalar}
part $q_w$, \revr{vector part $\vect{q}_v=[x,\,y,\,z]^\top$,} conjugate $\bm{q}^*$, and product $\otimes$;
the \emph{unit quaternions}
\[
  \mathbb{S}^3\triangleq\{\bm{q}\in\mathbb{H}:\|\bm{q}\|{=}1\}
\]
represent rotations, and the \emph{pure quaternions}
\[
  \mathbb{H}_p\triangleq\{\bm{q}\in\mathbb{H}:\mathrm{Re}(\bm{q}){=}0\}
\]
embed three-dimensional vectors via $\vect{a}\in\R^3\mapsto
[0,\,\vect{a}^\top]^\top$~\cite{selig2005}.
A dual quaternion
$\dq=\bm{q}_p+\varepsilon\,\bm{q}_d\in\Hd$ pairs a primal and a dual
quaternion through the dual unit $\varepsilon$
\revr{($\varepsilon^2{=}0$, $\varepsilon{\neq}0$)}, with product
\[
\dq_1\boxtimes\dq_2 =
\bm{q}_{p_1}\!\otimes \bm{q}_{p_2}
+
\varepsilon\,
\left(
\bm{q}_{p_1}\!\otimes \bm{q}_{d_2}
+
\bm{q}_{d_1}\!\otimes \bm{q}_{p_2}
\right)
\]
and conjugate
$\dq^*=\bm{q}_p^*+\varepsilon\,\bm{q}_d^*$.

\revr{The subset of unit dual quaternions is}
\begin{equation}
    \Sd \triangleq
    \left\{
    \dq\in\Hd :
    \|\bm{q}_p\|=1,\;
    \langle \bm{q}_p,\bm{q}_d\rangle=0
    \right\},
    \label{eq:unit_constraints}
\end{equation}
\revr{the double cover of $\SE$, used here to represent the quadrotor pose; the pure dual quaternions $\Hdp$, with $\mathrm{Re}(\bm{q}_p){=}\mathrm{Re}(\bm{q}_d){=}0$, represent dual twists and pose errors in Section~\ref{sec:mpcc}.}

\revr{\textit{Conventions.} Quaternions follow the Hamilton convention, scalar first; rotations act body-to-world; left-invariant pose errors
$\hat{\bm{q}}^e=\hat{\bm{q}}^{d*}\boxtimes\hat{\bm{q}}$ are expressed
in the desired body frame, the one attached to the reference pose and
distinct from the vehicle body frame of the prediction model, with the real scalar kept nonnegative
(jointly on real and dual parts) so that $\|\bm{\phi}\|\le\pi$.
For a rotation of angle $\theta$ about $\hat{\vect{n}}$, $q_w=\cos\tfrac{\theta}{2}$ and
$\vect{q}_v=\hat{\vect{n}}\sin\tfrac{\theta}{2}$~\cite{barfoot2017state}; inverting gives the
rotation vector
$\Log(\bm{q})=2\,\vect{q}_v\arctan(\|\vect{q}_v\|,q_w)/\|\vect{q}_v\|=\theta\hat{\vect{n}}$, smooth
for $\theta<\pi$ and not continuous at $\theta=\pi$, where $\pi\hat{\vect{n}}$ and
$-\pi\hat{\vect{n}}$ represent the same rotation.}

\subsection{Quadrotor Model}

Let $\vect{p}, \vect{v}, \vect{\omega}\in\R^3$ and
$\bm{q}_r\in\mathbb{S}^3$ denote the inertial position, inertial
velocity, angular velocity in the body frame, and attitude quaternion of a
quadrotor with mass $m$, respectively. We adopt a body-rate command
interface, where the onboard low-level rate controller tracks
$\vect{\omega}_{\mathrm{cmd}}$ according to a first-order lag with
time constant $\tau_{rc}$. The quadrotor model can be written as follows:
\begin{align}
    \dot{\vect{p}}      &= \vect{v},
                                        \label{eq:uav_p}\\
    \dot{\bm{q}}_r           &= \frac{1}{2}\,\bm{q}_r\otimes\vect{\omega},
                                        \label{eq:uav_q}\\
    \dot{\vect{v}}      &= \frac{f}{m}\,\bm{R}(\bm{q}_r)\,\vect{e}_3
                           + \vect{g},  \label{eq:uav_v}\\
    \dot{\vect{\omega}} &= \frac{1}{\tau_{rc}}
                           \bigl(\vect{\omega}_{\mathrm{cmd}}
                           -\vect{\omega}\bigr),
                                        \label{eq:uav_w}
\end{align}
with $f$ the collective thrust,
$\bm{R}(\bm{q}_r)\in\SO$ the rotation matrix of $\bm{q}_r$,
$\vect{e}_3=[0,0,1]^\top$, $\vect{g}=[0,0,-g]^\top$,
and control input
$\vect{u}=[f,\vect{\omega}_{\mathrm{cmd}}^\top]^\top\in\R^4$.

\subsection{Dual Quaternion Quadrotor Dynamics}
\label{sec:dq_state}

The position $\vect{p}$ and attitude $\bm{q}_r$ are unified
into a single unit dual quaternion
\begin{equation}
    \dq = \bm{q}_r + \varepsilon\,\bm{q}_d
    \;\in\;\Sd,
    \label{eq:dq_pose}
\end{equation}
where $\bm{q}_d\triangleq\tfrac{1}{2}\,\vect{p}\otimes \bm{q}_r$, with
$\vect{p}$ taken as a pure quaternion,
and $\bm{q}_r$ is the unit
quaternion. The position is recovered as
$\vect{p}=\bigl(2\,\bm{q}_d\otimes\bm{q}_r^{*}\bigr)\big|_{\mathrm{vec}}$
($|_{\mathrm{vec}}$: vector part).

With the dual-quaternion kinematics naturally in the body frame,
we use the velocity in the body frame
$\vect{v}_b=\bm{R}^{\top}(\bm{q}_r)\vect{v}$ and the dual twist
\[
\hat{\vect{\xi}}
=
\vect{\omega}
+
\varepsilon\,\vect{v}_b
\in\Hdp,
\]
where $\vect{\omega}$ and $\vect{v}_b$ are defined as pure quaternions. With this convention, the pose kinematics is written compactly as
\begin{equation}
    \dot{\dq}
    =
    \frac{1}{2}\,\dq\boxtimes\hat{\vect{\xi}}.
    \label{eq:dq_kinematics}
\end{equation}

Differentiating $\vect{v}_b$ and
substituting~\eqref{eq:uav_v}--\eqref{eq:uav_w} gives $\dot{\hat{\vect{\xi}}}=\hat{\vect{f}}+\hat{\vect{u}}$
with vector fields
\begin{align}
    \vect{f}_\omega &= -\tfrac{1}{\tau_{rc}}\,\vect{\omega},
    &
    \vect{f}_v &= \vect{v}_b\times\vect{\omega}
                   - \bm{R}^\top\!(\bm{q}_r)\,g\,\vect{e}_3,
    \label{eq:fv_fw}\\[4pt]
    \vect{u}_\omega &= \tfrac{1}{\tau_{rc}}\,\vect{\omega}_{\mathrm{cmd}},
    &
    \vect{u}_v &= \frac{f}{m}\,\vect{e}_3,
    \label{eq:uv_uw}
\end{align}
giving the pure dual quaternions
$\hat{\vect{f}}=\vect{f}_\omega+\varepsilon\,\vect{f}_v$
and
$\hat{\vect{u}}=\vect{u}_\omega+\varepsilon\,\vect{u}_v$.

The quadrotor dynamics can be compactly written using dual
quaternions as
\begin{equation}
\frac{d}{dt}
\begin{bmatrix}\dq\\[4pt]\hat{\vect{\xi}}\end{bmatrix}
=
\begin{bmatrix}
    \tfrac{1}{2}\,\dq\boxtimes\hat{\vect{\xi}}\\[6pt]
    \hat{\vect{f}}+\hat{\vect{u}}
\end{bmatrix},
\qquad
\dq\in\Sd,\;\;
\hat{\vect{\xi}},\;\hat{\vect{f}},\;\hat{\vect{u}}
\;\in\;\Hdp.
\label{eq:dynamics}
\end{equation}

The state and control are defined as follows:
\begin{equation}
    \vect{x}=[\dq,\;\hat{\vect{\xi}}]
    \in\Sd\times\Hdp,\quad
    \vect{u}=[f,\;\vect{\omega}_{\mathrm{cmd}}^\top]^\top\in\mathbb{R}^4.
    \label{eq:state_vector}
\end{equation}

\section{Reference Path Generation}
\label{sec:ref_traj}
MPCC uses a \emph{spatial} reference indexed by the path
arc-length $s$, so that the progress speed $v_s{=}\dot{s}$
becomes a decision
variable~\cite{lam2010mpcc,liniger2015optimization}.
The reference pose at progress~$s$ is
$\bm{\gamma}(s)=\bigl(\vect{p}^d(s),\;\bm{q}_r^d(s)\bigr)
\in\R^3\times\mathbb{S}^3$.
We build the reference in three stages: gate layout,
\revr{minimum-time} point-mass model (PMM) planning, and differential-flatness
lifting to an $\SE$ reference with arc-length reparametrization.
Each run comprises entry from hover, a low-speed warm-up, three timed
analysis laps (the only part entering the \revr{continuous metrics; failures count over the full run}), and an exit.

\paragraph{Gate layout}
The circuit has $N_g{=}8$ gates (center $\vect{p}_g$, normal
$\vect{n}_g$) in an
$11\!\times\!4.5\!\times\!3.65$\,m space, with a wall margin
confining the path to the inner corridor of
Fig.~\ref{fig:traj_racing}. Each gate has
aperture $R_{\rm gate}{=}0.50$\,m~\cite{kaufmann2023champion} and geometric
clearance $R_{\rm safe}{=}R_{\rm gate}{-}r_{\rm drone}{=}0.30$\,m, with
$r_{\rm drone}{=}0.20$\,m the propeller-tip radius.

\paragraph{\revr{Minimum-time} lap}
This stage shapes the racing path through the eight gates via an
offline minimum-time Optimal Control Problem (OCP) for a point-mass
model~\cite{foehn2021time,penicka2022minsnap}, with
state $[\vect{p}^d,\vect{v}^d]^{\!\top}$ and control acceleration $\vect{a}^d$:
\begin{subequations}
\label{eq:pmm_ocp}
\begin{align}
    \min_{T_\ell,\,\vect{a}^d(\cdot)}\quad & T_\ell\revr{+\lambda_a\!\int_0^{T_\ell}\!\|\vect{a}^d\|^2\,dt} \label{eq:pmm_obj}\\
    \mathrm{s.t.}\quad
        & \dot{\vect{p}}^d=\vect{v}^d,\qquad \dot{\vect{v}}^d=\vect{a}^d,
        \label{eq:pmm_dyn}\\
        & \vect{x}^d(0)=\vect{x}^d(T_\ell),
        \label{eq:pmm_per}\\
        & \|\vect{p}^d(t_k)-\vect{p}_{g,k}\|\le\varepsilon_g,
        \label{eq:pmm_center}\\
        & \vect{v}^d(t_k)\!\cdot\!\vect{n}_{g,k}\ge v_{\min},
        \label{eq:pmm_cross}\\
        & \|\vect{a}^d\|_2\le 5g,\qquad a^d_z\ge -0.9g,
        \label{eq:pmm_acc}
\end{align}
\end{subequations}
for $k{=}1,\dots,N_g$, where $\vect{x}^d{=}[\vect{p}^d,\vect{v}^d]^{\!\top}$
is the planner state, $T_\ell$ is the lap time, and $t_k$ is the time at which gate $k$
(center $\vect{p}_{g,k}$, normal $\vect{n}_{g,k}$) is crossed. The objective~\eqref{eq:pmm_obj} minimizes the lap time with a small control-effort regularization for smoothness \revr{($\lambda_a{=}10^{-6}$, SI)}. The periodicity constraint~\eqref{eq:pmm_per} yields a single repeatable lap, and~\eqref{eq:pmm_center}--\eqref{eq:pmm_cross} keep the path within $\varepsilon_g{=}5$\,cm of each gate center and require a forward crossing. In~\eqref{eq:pmm_acc} the bound $\|\vect{a}^d\|_2\le 5g$ limits the planned acceleration, while the thrust-positive bound $a^d_z\ge -0.9g$ \revr{keeps the required thrust pointing upward, so the} attitude reference \revr{stays} differentiable once the path is lifted to a full pose by differential flatness.

The $5g$ envelope exceeds the platform's maximum thrust
acceleration, so the point-mass lap cannot be tracked exactly by the real
vehicle. This is intentional: MPCC does not require a dynamically feasible
trajectory, only a continuously differentiable spatial
path~\cite{romero2022model}. The planner therefore defines the geometric path, the speed along it
being decided online under the true actuation limits.

A variable tube $d_{\rm tube}(s)$ is then constructed around the path: it
narrows at each gate and widens between gates, bounding the admissible
contour error along the path.

\paragraph{Arc-length reparametrization and flatness lifting}
The reference is resampled to arc-length
$s(t){=}\int_0^t\|\vect{v}^d\|d\tau$ and Gaussian-smoothed
($\sigma_a{=}25$\,ms) into a $C^2$, bounded-jerk profile
$\tilde{\vect{a}}^d(s)$. Differential
flatness~\cite{mellinger2011minimum}
then defines the thrust axis $\vect{b}_3^d$ and, with a
fixed-yaw convention $\psi^d{=}0$ standard in
racing~\cite{romero2022model,foehn2021time,penicka2022minsnap},
the remaining orthonormal columns of the rotation
$\bm{R}^d(s){=}[\vect{b}_1^d\,\vect{b}_2^d\,\vect{b}_3^d]\in\SO$:
\begin{equation}
    \vect{b}_3^d=\frac{\tilde{\vect{a}}^d+g\,\vect{e}_3}
                      {\|\tilde{\vect{a}}^d+g\,\vect{e}_3\|},
    \;\;\;
    \vect{b}_2^d=\frac{\vect{b}_3^d\times\vect{e}_1}
                      {\|\vect{b}_3^d\times\vect{e}_1\|},
    \;\;\;
    \vect{b}_1^d=\vect{b}_2^d\times\vect{b}_3^d,
    \label{eq:triad}
\end{equation}
with $\vect{e}_1$ the world $x$-axis.

$\bm{q}_r^d(s)$ is extracted from $\bm{R}^d$ with sign chosen
for continuity in $s$; the feedforward rate in the body frame is linear in
the progress speed,
\begin{equation}
    \vect{\omega}^d(s,v_s)=\bm{\omega}_0(s)\,v_s,\quad
    \bm{\omega}_0(s)\triangleq\Bigl({\bm{R}^d}^{\!\top}\tfrac{d\bm{R}^d}{ds}\Bigr)^{\!\vee}
    \in\R^3,
    \label{eq:omegad}
\end{equation}
where $(\cdot)^{\vee}:\so\!\to\!\R^3$ returns the vector of
a skew-symmetric matrix~\cite{lee2010geometric}.
A final Gaussian low-pass ($\sigma_q{=}85$\,ms) on
$\bm{q}_r^d(s)$ keeps $\bm{\omega}^d$ within the $20$\,rad/s
envelope. References are $C^2$ cubic B-splines evaluated symbolically inside
the OCP.
\revr{Since $\bm{R}^d(s)$ is lifted from the planner's acceleration profile it
is not recomputed as $v_s$ departs from the planned timing, only~\eqref{eq:omegad}
scaling with $v_s$; an attitude consistent with the timing would depend on
$(s,v_s,a_s)$. We retain the offline profile as a smooth $C^2$ heuristic soft
prior, weighted through the soft attitude term and never as a constraint.}

\section{DQ-MPCC}
\label{sec:mpcc}
\begin{figure}[!t]
    \centering
    \includegraphics[width=\columnwidth]{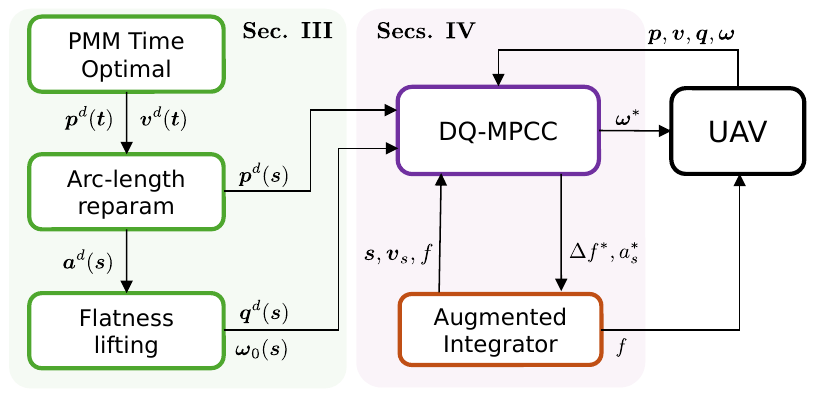}
    \caption{Control block diagram of the proposed \DQMPCC{},
    with shaded regions illustrating the subsections where each
    component is presented.}
    \label{fig:mpcc_blockdiagram}
\end{figure}

The architecture of \DQMPCC{} and its main components are shown in
Fig.~\ref{fig:mpcc_blockdiagram}. The offline pipeline provides the OCP with the B-splines
$\bm{\gamma}(s)$ and $\bm{\omega}_0(s)$.

\subsection{System Dynamics}
We augment the pose state with the arc-length progress pair
$(s,v_s)$, letting the controller adjust speed along the path, and
with the thrust $f$, whose rate $\Delta f$ is the smooth control,
\begin{equation}
    \dot{s}=v_s,\qquad \dot{v}_s=a_s,\qquad \dot{f}=\Delta f ,
    \label{eq:arc_dyn}
\end{equation}
with $a_s$ the progress acceleration. Stacked onto the
dual-quaternion state~\eqref{eq:state_vector}:
\begin{equation}
\begin{aligned}
    \vect{x}_{\mathrm{a}}
    &= \bigl[\dq,\;\hat{\vect{\xi}},\;s,\;v_s,\;f\bigr]
    \in\Sd\times\Hdp\times\R^3,\\
    \vect{u}_{\mathrm{a}}
    &= \bigl[\,\Delta f,\;\vect{\omega}_{\mathrm{cmd}}^\top,\;a_s\bigr]^\top
    \in\R^5 .
\end{aligned}
    \label{eq:aug_vectors}
\end{equation}

\subsection{Optimal Control Problem}
\label{sec:ocp}

Over a non-uniform grid of $N$ \revr{shooting intervals} spanning $T_p$,
\DQMPCC{} solves at every step:
\begin{subequations}
\label{eq:ocp}
\begin{align}
    \min_{\vect{x}_{\mathrm{a},0:N},\,
          \vect{u}_{\mathrm{a},0:N-1}}
        &\quad\sum_{k=0}^{N-1}\ell(\vect{x}_{\mathrm{a},k},
         \vect{u}_{\mathrm{a},k})
        +\ell_e(\vect{x}_{\mathrm{a},N})
        \label{eq:ocp_cost}\\
    \mathrm{s.t.}\quad
        &\vect{x}_{\mathrm{a},k+1}
         =\vect{F}(\vect{x}_{\mathrm{a},k},
         \vect{u}_{\mathrm{a},k}),
        \label{eq:ocp_dyn}\\
        &\vect{x}_{\mathrm{a},0}
         =\vect{x}_{\mathrm{a},\mathrm{init}},
        \label{eq:ocp_init}\\
        &\vect{g}(\vect{x}_{\mathrm{a},k},\vect{u}_{\mathrm{a},k})
         \le\vect{0}.
        \label{eq:ocp_g}
\end{align}
\end{subequations}
Here $\ell$ and $\ell_e$ are the stage and terminal costs,
$\vect{F}(\cdot)$ the ERK4-integrated augmented dynamics, and
$\vect{g}(\cdot)$ the input and state constraints.
Following~\cite{homburger2026millisecond}, the non-uniform grid
(Table~\ref{tab:params}) concentrates resolution near the current
state.

\subsection{Stage Cost}
\label{sec:stage_cost}

The stage cost combines the tangent-space pose error
($\bm{\phi}^e$, $\bm{\rho}_{\mathrm{lag}}$,
$\bm{\rho}_{\mathrm{cont}}$), control terms, and a progress
reward. With the body-rate error
$\tilde{\vect{\omega}}=\vect{\omega}^d(s,v_s)-\vect{\omega}_{\mathrm{cmd}}$
relative to the feedforward, the stage cost is defined as:
\begin{equation}
\begin{aligned}
    \ell(\vect{x}_{\mathrm{a}},\vect{u}_{\mathrm{a}})
        &= {\bm{\phi}^e}^\top\bm{Q}_\phi\,\bm{\phi}^e
         + \bm{\rho}_{\mathrm{lag}}^\top\bm{Q}_\ell\,
           \bm{\rho}_{\mathrm{lag}}
         + \bm{\rho}_{\mathrm{cont}}^\top\bm{Q}_c\,
           \bm{\rho}_{\mathrm{cont}}\\
        &\quad
         + W_{\Delta f}\,\Delta f^{\,2}
         + \tilde{\vect{\omega}}^\top
           \bm{W}_\omega\,\tilde{\vect{\omega}}
         - W_s\,v_s,
\end{aligned}
\label{eq:cost}
\end{equation}
where $\bm{Q}_\phi,\bm{Q}_\ell,\bm{Q}_c
\in\R^{3\times 3}$ are diagonal pose-error weights,
$W_{\Delta f}>0$ penalizes the thrust rate $\Delta f$ for
smoothness (cf.\ \eqref{eq:arc_dyn}),
$\bm{W}_\omega\in\R^{3\times 3}$ penalizes deviations from the
angular-rate feedforward
$\vect{\omega}^d(s,v_s)$~\eqref{eq:omegad},
and $W_s>0$ weights progress along the path. The terminal cost
$\ell_e$ retains only the pose-error terms of~\eqref{eq:cost}.

\subsection{Dual-Quaternion Pose Error}
\label{sec:pose_error}

Let $\dq^d(s)$ be the unit dual quaternion encoding the reference
pose $\bm{\gamma}(s)$ via~\eqref{eq:dq_pose}. The left-invariant
pose error
\begin{equation}
    \dq^e = \dq^{d\,*}(s)\boxtimes\dq
    \;\in\;\Sd
    \label{eq:dq_error}
\end{equation}
expresses the rotational and translational deviation jointly in the body frame of the desired
pose~\cite{bullo1995proportional}\revr{; its real part is the rotation error $\bm{q}_r^e\triangleq\bm{q}_r^{d*}\otimes\bm{q}_r$}; \revr{for $\dq,\dq^d\in\Sd$} it remains unit by group closure,
so the logarithmic map below is well defined\revr{; the predicted nodes leave $\Sd$ by the residuals of Section~\ref{sec:results}}.

\begin{figure}[!t]
    \centering
    \includegraphics[width=1\columnwidth]{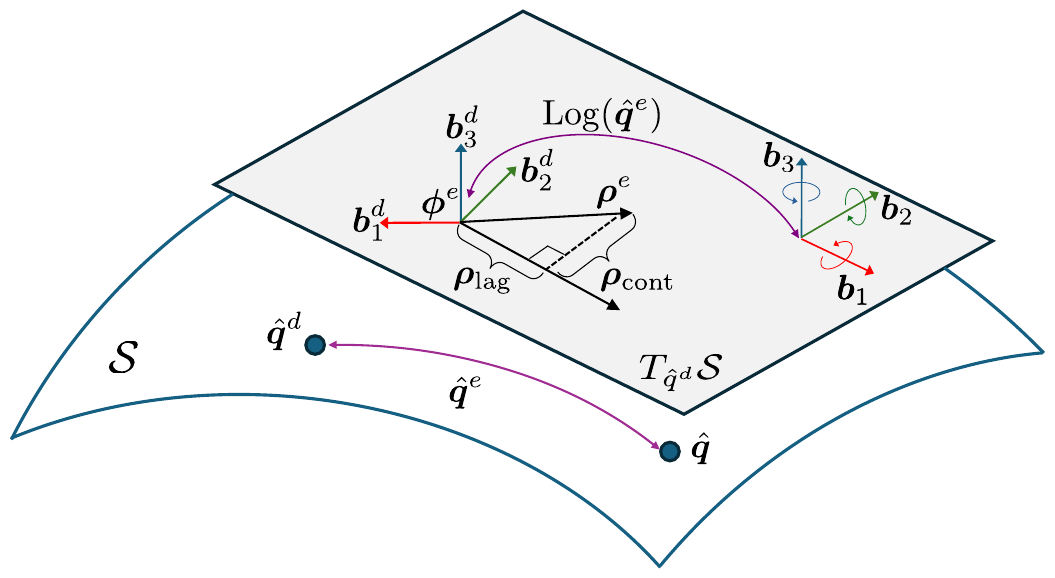}
    \caption{Dual-quaternion pose error: $\Log(\dq^e)$ lifts
    the error between $\dq$ and $\dq^d$ on $\Sd$ onto the
    tangent space at the desired pose, \revr{giving $\bm{\phi}^e$ and
    $\bm{\rho}^e$ in the desired body frame $\bm{b}^d_i$, the
    vehicle frame being $\bm{b}_i$; the translational coordinate is
    then decomposed along the path tangent into} $\bm{\rho}_{\mathrm{lag}}$ \revr{and}
    $\bm{\rho}_{\mathrm{cont}}$.}
    \label{fig:lie_decomposition}
\end{figure}

The logarithmic map projects $\dq^e$ onto $\Hdp$, the Lie
algebra of the unit dual quaternions; its six free parameters
are identified with $\se$~\cite{adorno2017fundamentals} and used
as the error coordinates throughout:
\begin{equation}
            \Log(\dq^e)
    = \begin{pmatrix}\bm{\phi}^e\\[2pt]
                     \bm{\rho}^e\end{pmatrix}\revr{\in\R^{6}},
    \label{eq:log_error}
\end{equation}
where \revr{$\bm{\phi}^e\triangleq\Log(\bm{q}_r^e)=\theta\hat{\vect{n}}$ is the rotation vector of the rotation error; DQ-NMPC~\cite{recalde2025dqnmpc} penalizes half of it, $\tfrac12\bm{\phi}$, and takes the position error itself as translational part, without the inverse left Jacobian of~\eqref{eq:rho_jinv}. The translational coordinate $\bm{\rho}^e$ follows from the position error in the desired body frame,
$\vect{t}^e\mathrel{\triangleq}\bigl(2\,\bm{q}_d^e\otimes \bm{q}_r^{e\,*}\bigr)\big|_{\mathrm{vec}}\revr{=\bm{R}^{d\top}(\vect{p}-\vect{p}^d)}$~\cite{adorno2017fundamentals},
through the inverse left Jacobian of $\SO$\revr{~\cite{selig2005,bullo2005geometric}}:}
\begin{equation}
        \bm{\rho}^e \mathrel{\revr{\triangleq}} \bm{J}_l^{-1}(\bm{\phi}^e)\,\vect{t}^e,
    \label{eq:rho_jinv}
\end{equation}
\revr{whose closed form, with $\bm{\phi}^e=\theta\hat{\vect{n}}$ the rotation vector and
$\alpha=\tfrac{\theta}{2}\cot\tfrac{\theta}{2}$,
is~\cite{barfoot2017state}
\begin{equation}
    \bm{J}_l^{-1}(\theta\hat{\vect{n}})
    = \alpha\,\bm{I}_3
    - \tfrac{\theta}{2}\,[\hat{\vect{n}}]_\times
    + (1-\alpha)\,\hat{\vect{n}}\hat{\vect{n}}^{\top},
    \label{eq:jl_inv_closed}
\end{equation}}
\revr{which the deployed controller approximates as}
$\bm{J}_l^{-1}\,{\approx}\,\bm{I}$, so that
$\bm{\rho}^e{=}\bm{R}^{d\top}(\vect{p}{-}\vect{p}^d)$ as
deployed\revr{: an identity approximation of the joint $\se$
pose-error coordinate about zero rotation error, not the exact logarithm.} \revr{Adopted as a reduced pose-error coordinate, it simplifies the translational mapping; Section~\ref{sec:results} quantifies its non-negligible racing-speed error and evaluates the exact coordinate at fixed gains.}

\subsection{Lag and Contouring Decomposition in $\se$}
\label{sec:lag_contour}

Projecting the error $\bm{\rho}^e$ \revr{in the desired body frame} onto the path tangent
$\vect{t}^d_b=\bm{R}^{d\top}(s)\,\vect{t}^d$, with
$\vect{t}^d$ the unit tangent of the reference path, decomposes it into
lag (along-path) and contouring (across-path) components
(Fig.~\ref{fig:lie_decomposition}):
\begin{equation}
    \bm{\rho}_{\mathrm{lag}}
    = \bigl(\vect{t}^d_b\,{\vect{t}^d_b}^\top\bigr)\,\bm{\rho}^e,
    \qquad
    \bm{\rho}_{\mathrm{cont}}
    = \bigl(\bm{I}_3-\vect{t}^d_b\,{\vect{t}^d_b}^\top\bigr)\,\bm{\rho}^e.
    \label{eq:lag_contour}
\end{equation}
With the rotation error $\bm{\phi}^e$, these form the pose tracking
error in the cost~\eqref{eq:cost}, expressed throughout in the desired
body frame.

\subsection{Constraints}
\label{sec:constraints}

The constraint function $\vect{g}(\vect{x}_{\mathrm{a}},
\vect{u}_{\mathrm{a}})\le\vect{0}$ in~\eqref{eq:ocp_g} collects the
actuator, progress, and corridor limits
\begin{subequations}
\label{eq:constraints}
\begin{align}
        &f_{\min}\le f\le f_{\max},
        \label{eq:ocp_thrust}\\
        &|\Delta f|\le\Delta f_{\max},
        \label{eq:ocp_df}\\
        &|\omega_{\mathrm{cmd},i}|\le\omega_{\max},
        \quad i=1,2,3,
        \label{eq:ocp_rate}\\
        &a_{s,\min}\le a_s\le a_{s,\max},
        \label{eq:ocp_vacc}\\
        &\|\bm{\rho}_{\mathrm{lag}}\|\le D_{\mathrm{lag}}+\sigma_\ell,
         \quad\sigma_\ell\ge 0,
        \label{eq:ocp_lag}\\
        &\|\bm{\rho}_{\mathrm{cont}}\|\le d_{\mathrm{tube}}+\sigma_c,
         \quad\sigma_c\ge 0,
        \label{eq:ocp_tube}\\
        &0\le s\le s_{\max}+s_m, \label{eq:ocp_arc}\\
        &0\le v_s\le v_{s,\max}.
        \label{eq:ocp_vs}
\end{align}
\end{subequations}
Constraints~\eqref{eq:ocp_lag}--\eqref{eq:ocp_tube} bound the lag
and contouring errors softly ($D_{\mathrm{lag}}$, tube radius $d_{\mathrm{tube}}(s)$; cost-penalized slacks)\revr{; $s_m$ is a terminal progress margin}.
The norm constraint in~\eqref{eq:unit_constraints} is
maintained by a Baumgarte-type correction
$K_q(1-\|\bm{q}_r\|)\,\bm{q}_r$, $K_q{=}10$\,s$^{-1}$, added to $\dot{\bm{q}}_r$ inside the
integrator~\cite{baumgarte1972}; the orthogonality constraint is
re-established at every control iteration, where $\dq$ is rebuilt
from the estimated pose via~\eqref{eq:dq_pose}.

Physical states come from
the state estimator; virtual states $(s,v_s,f)$ are
Euler-integrated locally at the loop period $\dt$ from
$\vect{u}_{\mathrm{a},0}^{*}$,
\begin{equation}
\begin{aligned}
    s_{k+1}     &= s_k     + v_{s,k}\,\dt,\\
    v_{s,k+1}   &= v_{s,k} + a_s^{*}\,\dt,\\
    f_{k+1}     &= f_k     + \Delta f^{*}\,\dt,
\end{aligned}
\label{eq:virtual_integration}
\end{equation}
with drift absorbed by re-initialization~\eqref{eq:ocp_init} each
iteration.

\section{Experimental Setup}
\label{sec:exp_setup}

\textbf{Numerical implementation.}
The OCP is solved with acados~\cite{verschueren2021acados} via
Sequential Quadratic Programming Real-Time
Iterations (SQP-RTI)\revr{~\cite{gros2020rti}};
simulations use MuJoCo with the parameters of
Table~\ref{tab:params}.
Inequality constraints are slack-relaxed to keep the QP feasible.
All SiL evaluations run on the same Jetson Orin~NX used in
the real-world experiments.

\begin{table}[!t]
\centering
\caption{MPCC vs DQ control formulations.
         Quat: unit quaternion; DQ: unit dual quaternion;
         PF: path following; TK: tracking.
         $^\dagger$Non-uniform grid (Section~\ref{sec:ocp}).
         \revr{$\se$ is the formulation of~\eqref{eq:rho_jinv}, deployed with $\bm{J}_l^{-1}{\approx}\bm{I}$.}}
\label{tab:comparison}
\setlength{\tabcolsep}{2pt}
\resizebox{\columnwidth}{!}{
\begin{tabular}{lcccccc}
\toprule
Feature
  & \cite{liniger2015optimization}
  & \cite{cmpcc2021}
  & \cite{romero2022model}
  & \cite{mpccpp2024}
  & \cite{recalde2025dqnmpc}
  & \textbf{Ours} \\
\midrule
Control type
  & PF
  & PF
  & PF
  & PF
  & TK
  & PF \\
Dynamics model
  & Bicycle
  & Point mass
  & Full 6-DoF
  & Full 6-DoF
  & Full 6-DoF
  & Full 6-DoF \\
Pose error space
  & $\mathbb{R}^2$
  & $\mathbb{R}^3$
  & $\mathbb{R}^3 \!\times\! \so$
  & $\mathbb{R}^3 \!\times\! \so$
  & $\se$
  & $\se$ \\
Rotation repr.
  & ---
  & ---
  & Quat
  & Quat
  & DQ
  & DQ \\
\revr{Unified coords., body frame}
  & ---
  & ---
  & \ding{55}
  & \ding{55}
  & \ding{51}
  & \ding{51} \\
Lag and cont.\ in $\se$
  & \ding{55}
  & \ding{55}
  & \ding{55}
  & \ding{55}
  & ---
  & \ding{51} \\
Progress dynamics
  & $\dot{s}{=}v_s$
  & $\ddot{s}{=}a_s$
  & $\ddot{s}{=}a_s$
  & $\ddot{s}{=}a_s$
  & ---
  & $\ddot{s}{=}a_s$ \\
Safety constraints
  & Track bounds
  & Corridor
  & ---
  & Gate avoid.
  & ---
  & Soft lag, tube \\
Horizon $N$ / $T_p$
  & $40$ / $0.8$\,s
  & $20$ / $1.0$\,s
  & $20$ / $0.8$\,s
  & $20$ / $0.8$\,s
  & ---\,/ $1.5$\,s
  & $31$ / $1.48$\,s$^\dagger$ \\
\revr{Solved onboard}
  & ---
  & \ding{51}
  & \ding{55}
  & \ding{55}
  & \ding{51}
  & \ding{51} \\
\bottomrule
\end{tabular}}
\end{table}

\textbf{Baseline.}\label{sec:baseline} \revr{The baseline matches \DQMPCC{}
in solver, horizon, constraints, reference, actuation, and cost structure,
but differs in state and pose-error coordinates:} it propagates the inertial body-rate
model~\eqref{eq:uav_p}--\eqref{eq:uav_w}, augmented with the same arc-length and
thrust states~\eqref{eq:arc_dyn}, giving the state
$\vect{x}_{\rm e}{=}(\vect{p},\vect{v},\bm{q}_r,\vect{\omega},s,v_s,f)$ in place of
the dual-quaternion state in the body frame~\eqref{eq:aug_vectors}, and it uses the
standard $\R^3{\times}\so$ error decomposition in place of the $\se$ one. The
inertial position error $\vect{e}_p=\vect{p}^d(s)-\vect{p}$ decouples into
along-path and across-path components,
\[
  \vect{e}_\ell=(\vect{t}^{d\,\top}\vect{e}_p)\,\vect{t}^d,
  \qquad
  \vect{e}_c=(\bm{I}_3-\vect{t}^d\,\vect{t}^{d\,\top})\,\vect{e}_p,
\]
with attitude error
$\bm{\eta}=\Log(\bm{q}_r^{*}\otimes\bm{q}_r^d)\in\so$ \revr{in the body frame}; the baseline
cost is~\eqref{eq:cost} \revr{with $\bm{\eta}$, $\vect{e}_\ell$, and $\vect{e}_c$ in place of
$\bm{\phi}^e$, $\bm{\rho}_{\mathrm{lag}}$, and $\bm{\rho}_{\mathrm{cont}}$}, with the
control and progress terms unchanged.

\revr{For the deployed approximation,} this cost \revr{has the same term structure as} the \DQMPCC{} one: $\bm{R}^d$ being
orthogonal,
$\bm{\rho}_{\mathrm{lag}}{=}{-}\bm{R}^{d\top}\vect{e}_\ell$,
$\bm{\rho}_{\mathrm{cont}}{=}{-}\bm{R}^{d\top}\vect{e}_c$,
\revr{so the translational errors share the same norm. The two error quaternions are conjugates, so the attitude errors are the same rotation vector up to sign, $\bm{\phi}^e=-\bm{\eta}$, and the attitude penalties coincide under matched weights $\bm{Q}_\phi=\bm{Q}_q$. With the isotropic pose-error weights of Section~\ref{sec:pareto_tuning} the cost shape therefore does not differ between the controllers.} \revr{Equality of
the instantaneous error norms does not make the two nonlinear
programs equivalent: the errors are functions of different state
coordinates, so the second derivatives of the two stage costs
differ, and the prediction models propagate those coordinates
differently over the horizon.} \revr{Performance differences therefore cannot reflect missing physical coupling; they can arise from the coordinates in which the problem is stated and from the independently tuned gains, which remain a confound (Section~\ref{sec:results}).}
Comparing directly to published racing
MPCCs~\cite{romero2022model,cmpcc2021,mpccpp2024} would confound
solver, tuning, actuation, and platform with the $\SE$ effect\revr{, the two with the full
model being solved off board}; Table~\ref{tab:comparison} positions both formulations instead.

\textbf{Metrics.} $T_{\min}$/$T_{\rm avg}$: minimum/average lap
time over the $N_\ell$ analysis laps; $\delta$: in-plane gate-crossing offset from the gate center, \revr{with maximum $\delta_{\max}$ and average $\delta_{\rm avg}$ over the crossings of a run}. \revr{\emph{Crossing-margin compliance} is $\delta{\le}R_{\rm safe}$ at every crossing; this event metric does not guarantee collision-free traversal elsewhere.} $e_c$, $e_\ell$:
contour and lag RMSE from the same inertial path projection
(independent of the representation); $L_p$: per-lap path length;
$|\vect{v}|_{\rm max}$, $|\vect{a}|_{p99}$,
$|\bm{\omega}|_{\rm rms}$: maximum speed, $99$th-percentile
acceleration, RMS body rate over the racing laps (odometry and IMU).

\begin{table}[!t]
\centering
\caption{System and OCP Parameters.}
\label{tab:params}
\setlength{\tabcolsep}{1.5pt}
\footnotesize
\begin{tabular}{ll ll}
\toprule
\multicolumn{2}{c}{Platform} & \multicolumn{2}{c}{Solver / OCP} \\
\cmidrule(lr){1-2}\cmidrule(lr){3-4}
$m$            & $1.08$\,kg        & solver        & acados SQP-RTI \\
$\tau_{rc}$    & $30$\,ms          & integrator    & ERK4 \\
$f$            & $[0,49.05]$\,N    & horizon $N$   & $31$ \\
$|\Delta f|$   & $\le 500$\,N/s    & $T_p$         & $1.48$\,s \\
$|\omega_{\mathrm{cmd},i}|$ & $\le 20$\,rad/s & loop rate & $100$\,Hz \\
$|a_s|$        & $\le 5g$          & regularization & PROJECT$+$LM ($10^{-2}$) \\
$D_{\rm lag}$  & $0.5$\,m          & grid [ms]     & $10{\times}20{+}10{\times}40{+}11{\times}80$ \\
$s_{\max}$     & ${\approx}118$\,m & QP solver & {PARTIAL\_CONDENSING\_HPIPM}  \\
$v_{s,\max}$   & ${\approx}14.65$\,m/s & Hessian approx. & Gauss Newton \\

\bottomrule
\end{tabular}
\end{table}

\subsection{Multi-Objective Tuning}
\label{sec:pareto_tuning}

Both controllers are tuned offline with
Optuna~\cite{akiba2019optuna}: a broad multi-objective
NSGA-III~\cite{deb2014nsgaiii} exploration, a TPE refinement,
and a final evaluation under matched progress-speed bounds
$v_{s,\max}$, so that performance differences are not driven by
different speed limits. Each stage uses $100$ trials per controller
and $n_{\rm seeds}=3$ MuJoCo seeds.

The optimized gain vector
$\bm{\theta}=\bigl[q_\phi,\,q_c,\,q_\ell,\,w_\omega,\,w_s,\,
v_{s,\max}\bigr]$
maps to the cost weights of~\eqref{eq:cost} as
$\bm{Q}_\phi = q_\phi\bm{I}_3$, $\bm{Q}_c = q_c\bm{I}_3$,
$\bm{Q}_\ell = q_\ell\bm{I}_3$, $\bm{W}_\omega = w_\omega\bm{I}_3$, and
$W_s = w_s$. \revr{$W_{\Delta f}$ is fixed at $10^{-3}$, not searched.}

\textbf{Racing objectives.}
Gate crossings are sign changes of
$\bm{n}_k^\top(\bm{p}-\bm{g}_k)$ within $2R_{\rm gate}$ of the gate
center; after entry, the next $N_\ell{+}1$ crossings of $G_0$ define
the racing laps. The tuning problem is
formulated as the bi-objective optimization
\begin{equation}
  \min_{\bm{\theta}}
  \left(
  T_{\min}(\bm{\theta}) + \mathcal{P}(\bm{\theta}),\;
  \delta_{\max}^{\rm worst}(\bm{\theta})
  \right),
\end{equation}
where $\mathcal{P}$ penalizes gate safety, completion, body rate, and jerk violations, applied only to the lap-time objective,
\revr{keeping the worst-case gate offset unpenalized and the trade-off between speed and gate-crossing margin explicit}.

\textbf{Gain selection.}
Per controller, \revr{under identical criteria,} the \revr{ten fastest configurations whose worst tuning offset stays within $R_{\rm safe}$} are re-evaluated over five full-path SiL runs on the onboard computer \revr{with physically solid gates. Candidates completing all five are retained, and the fastest retained candidate is selected. Because a five-run screening is underpowered, deployment further requires $\delta_{\max}\le R_{\rm safe}$ across the $n{=}100$ evaluation of Section~\ref{sec:results}. The baseline's fastest five-run candidate ($2.86$\,s) exceeds $R_{\rm safe}$ in $10$ of its $93$ completed runs and is rejected; the deployed baseline satisfies it. A re-evaluation under one protocol ($n{=}30$ screening, $n{=}100$ validation) confirms it. All seven non-dominated baseline configurations faster than the deployed gains exceed $R_{\rm safe}$: six already at $n{=}30$ ($34.0$--$46.1$\,cm), hence also at $n{=}100$ since the criterion is a maximum over runs, and the seventh at $n{=}100$ ($48.8$\,cm). That seventh is also the only examined baseline configuration striking a gate inside the racing laps ($7/100$, against none for either deployed pair). $R_{\rm safe}$ is a geometric clearance, not an adjustable tolerance. The deployed \DQMPCC{} gains, two
further front configurations, and the exact-$\bm{J}_l^{-1}$ ablation at the deployed gains all pass both stages.}
Implementation and results will be released publicly upon acceptance.

\section{Simulation Results}
\label{sec:results}

We evaluate the deployed gains in SiL over $n{=}100$ Monte
Carlo seeds (\revr{pass-through gates;} injected odometry noise\revr{, $\sigma{=}1$\,cm, $0.5^{\circ}$, $0.01$\,m/s, $0.035$\,rad/s}; randomized warm-start;
Section~\ref{sec:pareto_tuning})\revr{; Table~\ref{tab:sil_racing}
reports medians over completed runs: of $100$ attempts, $8$ and $10$ are lost to a faulty simulator reset and $92$ and $90$ complete the laps; the campaign with solid gates below was run until $100$ started correctly}.

\begin{table}[!t]
\centering
\caption{SiL racing metrics, $n{=}100$ Monte Carlo seeds;
median$\,\pm\,$std over completed runs;
\textbf{bold} = better \revr{of the two deployed controllers;
$^{\dagger}$ = same-gains ablation with the exact coordinate~\eqref{eq:rho_jinv} (Section~\ref{sec:results}); solve times: overhead vs.\ \DQMPCC{} measured side by side}}
\label{tab:sil_racing}
\setlength{\tabcolsep}{4pt}
\scriptsize
\setlength{\tabcolsep}{1.5pt}
\begin{tabular}{l c c c}
\toprule
Metric & MPCC & \DQMPCC{} & \revr{exact $\bm{J}_l^{-1}$$^{\dagger}$} \\
\midrule
$T_{\min}$ [s]             & $3.457\,{\pm}\,0.010$ & $\mathbf{3.226\,{\pm}\,0.007}$ & \revr{$3.141\,{\pm}\,0.012$}  \\
$T_{\rm avg}$ [s]         & $3.478\,{\pm}\,0.006$ & $\mathbf{3.252\,{\pm}\,0.004}$ & \revr{$3.164\,{\pm}\,0.011$}  \\
\midrule
$\delta_{\max}$ [cm] & $\mathbf{\revr{17.88}\,{\pm}\,0.62}$ & $\revr{24.53}\,{\pm}\,0.73$ & \revr{$24.55\,{\pm}\,0.80$}  \\
$\delta_{\rm avg}$ [cm]   & $\mathbf{12.01\,{\pm}\,0.20}$ & $15.85\,{\pm}\,0.20$ & \revr{$15.44\,{\pm}\,0.44$}  \\
$e_c$ RMSE [cm]            & $\mathbf{5.23\,{\pm}\,0.14}$ & $12.70\,{\pm}\,0.11$ & \revr{$12.04\,{\pm}\,0.22$}  \\
$e_\ell$ RMSE [cm]            & $\mathbf{1.44\,{\pm}\,0.02}$ & $1.60\,{\pm}\,0.04$ & \revr{$1.57\,{\pm}\,0.04$}  \\
\midrule
$|\bm v|_{\rm max}$ [m/s]  & $\mathbf{11.19}$ & $11.00$ & \revr{$11.09$}  \\
$|\bm a|_{p99}$ [m/s$^2$]  & $29.7\,{\pm}\,0.4$ & $29.9\,{\pm}\,0.4$ & \revr{$29.7\,{\pm}\,1.0$}  \\
$|\bm\omega|_{\rm rms}$ [rad/s] & $5.28\,{\pm}\,0.03$ & $5.68\,{\pm}\,0.02$ & \revr{$5.73\,{\pm}\,0.11$}  \\
\midrule
$L_p$ [m/lap]              & $22.44\,{\pm}\,0.06$ & $\mathbf{21.80\,{\pm}\,0.06}$ & \revr{$21.80\,{\pm}\,0.07$}  \\
\midrule
solve median [ms]          & $\mathbf{5.88}$ & $7.60$ & \revr{$+0.2$}  \\
solve $p_{99}$ [ms]        & $\mathbf{7.64}$ & \revr{$9.63$} & \revr{$+0.2$}  \\
\midrule
\revr{crashes with solid gates ($n{=}100$)} & \revr{$17/100$} & \revr{$\mathbf{2/100}$} & \revr{$0/100$}  \\
\bottomrule
\end{tabular}
\end{table}

\textbf{Lap time.} \DQMPCC{} reduces $T_{\min}$ by $6.7\%$
($3.46\!\to\!3.23$\,s) along a $2.9\%$ shorter racing path
($L_p{=}21.8$ vs.\ $22.4$\,m/lap). The reduction comes with a similar $|\bm a|_{p99}$, within the spread (\revr{$29.7$ vs.\ $29.9$}\,m/s$^2$), and a slightly higher body rate.

\revr{\textbf{Scope of the comparison.} The two prediction models describe the same physical system; with consistently transformed objectives and constraints, exact integration, and full convergence, the coordinate descriptions are equivalent (Section~\ref{sec:baseline}). A single real-time iteration linearizes a different expression in each formulation: the attitude enters the translational dynamics with the thrust prefactor $f/m$ in~\eqref{eq:uav_v} and with the constant $\|\vect{g}\|$ in~\eqref{eq:fv_fw}, the thrust-to-weight ratio, up to $3$ in the racing laps. With solver, horizon, constraints, reference, and cost shape matched, coordinate-dependent local models and independently tuned gains remain different (contour-to-lag ratio: $287$ versus $1.3$). The block above exemplifies one such local-linearization difference but does not isolate representation from tuning. On the tuning fronts the baseline reaches lower lap times at every matched level, though every such configuration fails the deployment rule, and exchanging the gain sets leaves neither controller deployable, each set being the operating point of its own formulation.}

\textbf{\revr{Runtime and constraints.}} The dual-quaternion state adds $\sim1.7$\,ms of solve time; both
controllers \revr{run without solver failures (one baseline cycle excepted)}.
\revr{Over the racing laps, $10$\,ms solve overruns are rare ($0.01\%$ MPCC,
$0.45\%$ \DQMPCC{}; the previous command persists); neither saturates its authority in simulation ($p_{99}$: thrust ${\le}64\%$ of $f_{\max}$, body rate ${\le}76\%$ of $\omega_{\max}$). Over the racing laps the tube is never violated and the lag bound is exceeded in $4.1\%$ and $0.003\%$ of samples (worst $53$ and $4$\,cm), the baseline therefore running with its lag bound relaxed over part of the laps; with transients the tube is violated in $0.59\%$ and $0.39\%$ (worst $96$ and $24$\,cm). The predicted states are not projected: in ten replayed SiL runs, over the racing laps they stay within
$1.3\times10^{-2}$ of unit norm and $2.4\times10^{-2}$ of dual orthogonality, with the first node rebuilt every cycle. Solve times are measured on the deployment computer against the simulated plant, at the clock and solver build of the flights.}

\textbf{Contour precision.} In simulation the baseline tracks the path
more tightly, in contour RMSE ($5.2$ vs.\ $12.7$\,cm) and crossing offset
($\delta_{\max}{=}\revr{17.9}$ vs.\ $\revr{24.5}$\,cm): \DQMPCC{} is faster
but not uniformly more accurate, every crossing still inside $R_{\rm safe}$.

\revr{\textbf{Exact $\bm{J}_l^{-1}$ ablation.} The approximation of
Section~\ref{sec:pose_error} is not negligible here: attitude errors to the
nearest path point average ${\approx}16^{\circ}$ ($p_{99}$ $52$--$56^{\circ}$)
in simulation and flight, for which
$\|\bm{J}_l^{-1}(\bm{\phi}^e)\vect{t}^e-\vect{t}^e\|$ averages $11\%$ of
$\|\vect{t}^e\|$ ($p_{99}$ $42$--$46\%$). At the deployed gains the exact
$\bm{J}_l^{-1}$ gives a lower median lap time, similar tracking metrics, zero rather than $2/100$ crashes, adds approximately $0.2$\,ms of solve time, and leaves $5/100$ pass-through runs unfinished. This fixed-gain sensitivity test uses gains tuned for the deployed approximation, not an independently retuned comparison; the main comparison uses the slower deployed variant.}

\revr{\textbf{Solid-gate evaluation.}} \revr{Crossing offsets alone understate the solid-gate crash rates of Table~\ref{tab:sil_racing}: the
deployed baseline stays inside $R_{\rm safe}$ at the measured
crossings, yet contacts gates in the deceleration transient (all $17$ after the last lap; \DQMPCC{}'s two in the entry); runs and attempts count whole, both transients being flown by the racing controller.}

\begin{figure*}[!t]
  \centering
  \includegraphics[width=0.95\textwidth]{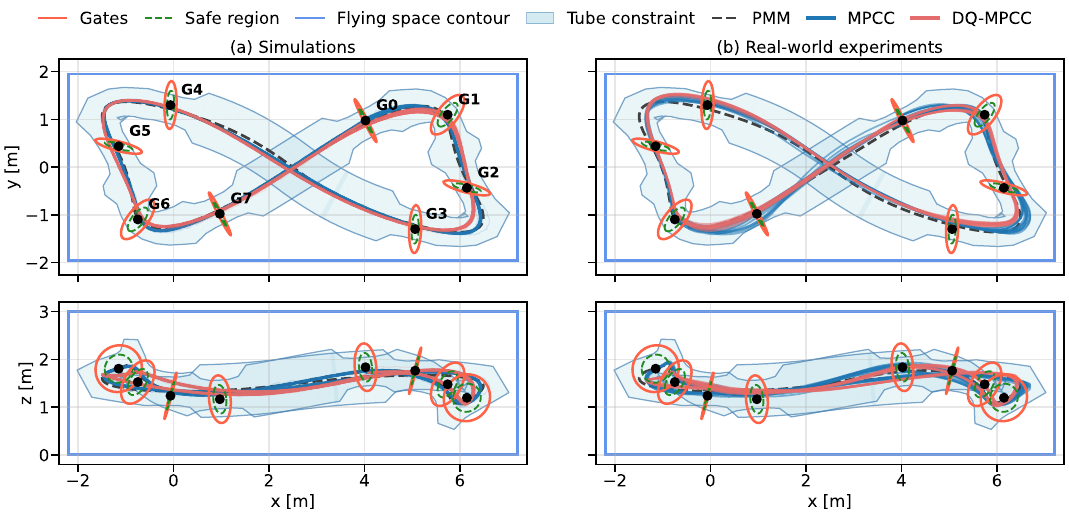}
    \caption{Racing laps for MPCC vs \DQMPCC{} over the contour tube and gate
    rings, $XY$ (top) and $XZ$ (bottom): (a) software-in-the-loop, $92$ and $90$ completed runs of the $n{=}100$ campaigns; (b) Real-world experiments, six flights per controller.}
  \label{fig:traj_racing}
\end{figure*}
\section{Real-World Experiments}
\label{sec:hardware}

\begin{figure}[!t]
  \centering
  \includegraphics[width=0.9\columnwidth]{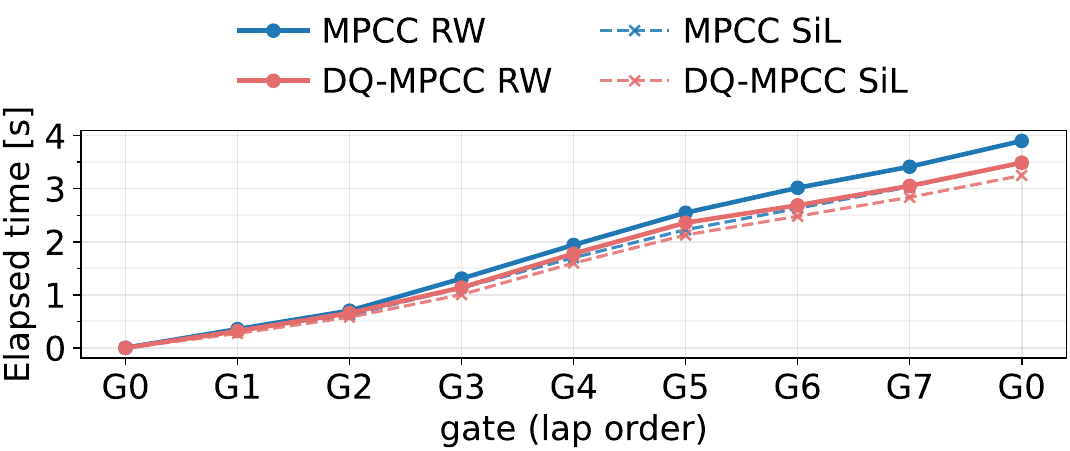}
  \caption{Per-gate lap-time build-up (median over racing laps). \DQMPCC{}'s
    lead over the baseline grows over the second half of the lap and is
    larger in the real world than in simulation.}
  \label{fig:laptime}
\end{figure}

\begin{figure}[!t]
  \centering
  \includegraphics[width=0.9\columnwidth]{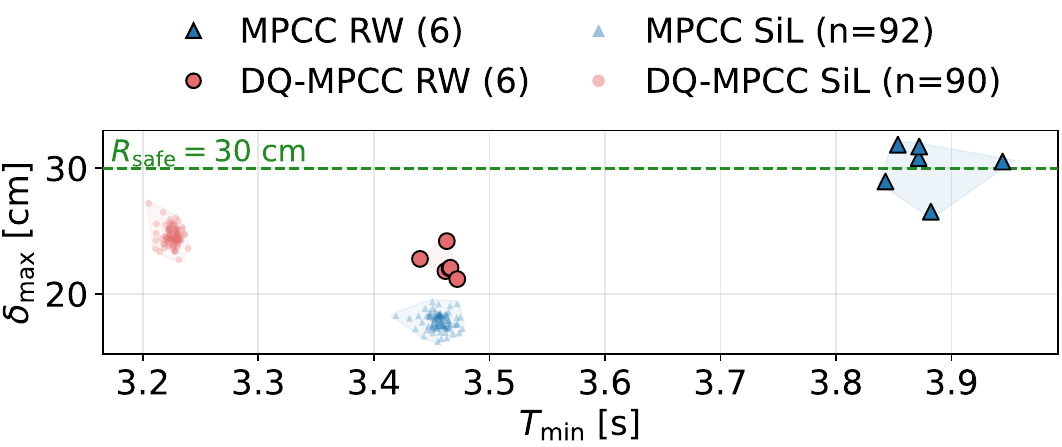}
  \caption{Speed--crossing-margin plane: SiL clouds and RW flights (convex hulls). From simulation to the real world the baseline shifts across
    $R_{\rm safe}$ whereas \DQMPCC{} stays inside it.}
  \label{fig:pareto}
\end{figure}

The platform is a custom quadrotor with an onboard NVIDIA
Jetson Orin~NX running the control optimization at $100$\,Hz, a
Betaflight autopilot with \revr{incremental nonlinear dynamic inversion (INDI)} body-rate control, and external
motion-capture state at $240$\,Hz.
Both controllers fly the circuit of Fig.~\ref{fig:portada}
with the SiL gains \revr{and no retuning}.
\revr{SiL lap-time, tracking, and offset metrics (Figs.~\ref{fig:laptime}, \ref{fig:pareto}) use non-colliding rings, since offsets beyond $R_{\rm safe}$ are observable only without contact; crash rates with solid gates come from separate contact campaigns.} Table~\ref{tab:accounting} reports the trial
accounting\revr{: in-air racing attempts continued until six flights per controller completed}
the three analysis laps ($18$ laps per controller; Fig.~\ref{fig:traj_racing}) from a
fixed warm-start hover. MPCC failed five times at racing speed and
\DQMPCC{} once below racing altitude\revr{, each after a solver failure reported by the controller}.

\begin{table}[!t]
\centering
\caption{Real-world trial accounting (racing attempts in flight).}
\label{tab:accounting}
\setlength{\tabcolsep}{4pt}
\footnotesize
\begin{tabular}{l c c c c}
\toprule
 & attempts & compl.\ ($3$ laps) & \revr{failed} & \revr{margin comp.} \\
\midrule
MPCC      & $11$ & $6$ & $5$ & $2/6$ \\
\DQMPCC{} & $7$  & $6$ & $\mathbf{1}$ & $\mathbf{6/6}$ \\
\bottomrule
\end{tabular}
\end{table}

\begin{table}[!t]
\centering
\caption{RW racing metrics, median$\,\pm\,$std over the six completed
flights per controller. $\Delta$ vs MPCC; \textbf{bold} = better.}
\label{tab:hw_racing}
\setlength{\tabcolsep}{4pt}
\footnotesize
\begin{tabular}{l c c c}
\toprule
Metric & MPCC & \DQMPCC{} & $\Delta$ \\
\midrule
$T_{\min}$ [s]              & $3.872\,{\pm}\,0.032$ & $\mathbf{3.464\,{\pm}\,0.010}$ & $\mathbf{-10.5\%}$ \\
$T_{\rm avg}$ [s]          & $3.906\,{\pm}\,0.030$ & $\mathbf{3.493\,{\pm}\,0.009}$ & $\mathbf{-10.6\%}$ \\
\midrule
$\delta_{\max}$ [cm]        & $30.7\,{\pm}\,1.8$ & $\mathbf{22.1\,{\pm}\,1.0}$ & $\mathbf{-28.1\%}$ \\
$\delta_{\rm avg}$ [cm]    & $\mathbf{13.0\,{\pm}\,0.8}$ & $14.7\,{\pm}\,0.4$ & $+13.1\%$ \\
$e_c$ RMSE [cm]             & $\mathbf{11.5\,{\pm}\,0.9}$ & $14.0\,{\pm}\,0.4$ & $+21.3\%$ \\
$e_\ell$ RMSE [cm]             & $\mathbf{4.59\,{\pm}\,0.01}$ & $4.82\,{\pm}\,0.01$ & $+5.1\%$ \\
\midrule
$|\bm v|_{\rm max}$ [m/s]  & $\mathbf{10.45\,{\pm}\,0.10}$ & $10.40\,{\pm}\,0.06$ & $-0.5\%$ \\
$|\bm a|_{p99}$ [m/s$^2$]   & $34.9\,{\pm}\,2.5$ & $28.7\,{\pm}\,0.2$ & $-18.0\%$ \\
$|\bm\omega|_{\rm rms}$ [rad/s] & $6.66\,{\pm}\,0.26$ & $6.47\,{\pm}\,0.08$ & $-2.9\%$ \\
\midrule
$L_p$ [m/lap]              & $21.28\,{\pm}\,0.08$ & $\mathbf{20.75\,{\pm}\,0.05}$ & $\mathbf{-2.5\%}$ \\
\bottomrule
\end{tabular}
\end{table}

\begin{figure}[!t]
  \centering
  \includegraphics[width=0.9\columnwidth]{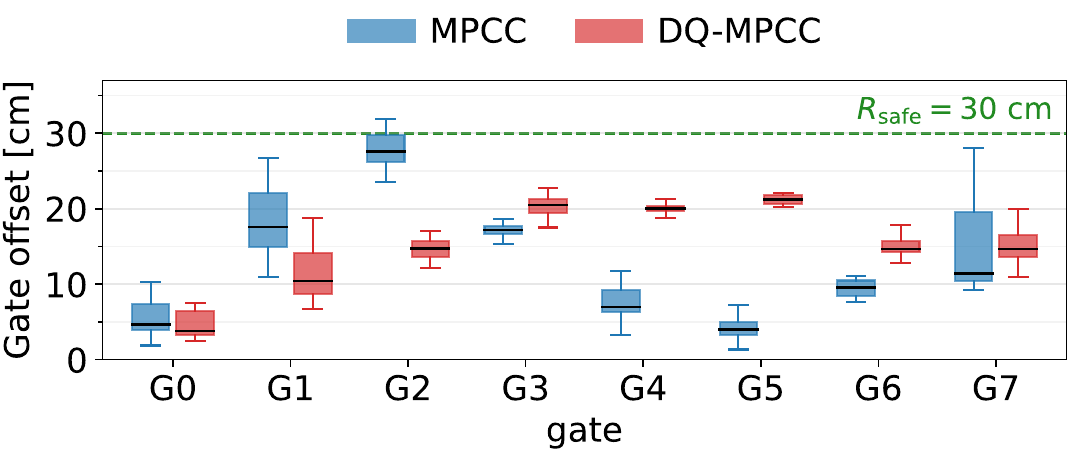}
  \caption{Gate-crossing offset at each gate over the RW racing laps. \DQMPCC{} stays inside
    $R_{\rm safe}$ at every gate; the baseline spreads to the boundary and
    crosses it at its worst gate, $G_2$.}
  \label{fig:miss_gate}
\end{figure}

The real world preserves the lap-time ranking ($-10.5\%$, larger
than SiL; Fig.~\ref{fig:laptime}) but the worst-case gate offset
\emph{reverses}: \DQMPCC{} keeps every gate inside
$R_{\rm safe}$ ($6/6$ flights, $\delta_{\max}{=}22.1$\,cm),
whereas the MPCC baseline, tighter in simulation, now
deviates by $30.7$\,cm, failing \revr{crossing-margin compliance} on $4$ of $6$
flights (Fig.~\ref{fig:miss_gate}). The reversal is confined to the
worst gate: the baseline keeps lower \emph{average} offset and
contour error (Table~\ref{tab:hw_racing}).

\revr{The sim-to-real change in gate offset is accompanied by a different change in the actuation envelope:} under real-world mismatch and attitude lag, the baseline's $|\bm a|_{p99}$
increases by $17.5\%$
sim-to-real (to $34.9$\,m/s$^2$), whereas \revr{\DQMPCC{}'s decreases by $4.1\%$} ($28.7$\,m/s$^2$). \revr{In the real world the baseline's larger corrections coincide with
crossings outside $R_{\rm safe}$, latency and unmodeled dynamics acting
jointly. The body-frame formulation carries a countervailing sensitivity: an attitude
estimate error $\delta\bm{\theta}$ perturbs $\vect{v}_b{=}\bm{R}^{\top}\vect{v}$
by at most $\|\vect{v}\|\,\|\delta\bm{\theta}\|$, a sensitivity the
inertial-coordinate baseline does not carry.}

\begin{table}[!t]
\centering
\caption{Sim-to-real gap (no retuning): per-controller RW-vs-SiL change.
\textbf{bold}: smaller absolute RW-vs-SiL change. \revr{Relative changes use unrounded medians.}}
\label{tab:sim2real}
\setlength{\tabcolsep}{3pt}
\footnotesize
\begin{tabular}{l c c c c c c}
\toprule
Ctrl. & $\Delta T_{\min}$ & $\Delta\delta_{\max}$ & $\Delta e_c$
      & $\Delta|\bm a|_{p99}$ & $\Delta\omega_{\rm rms}$ & $\Delta L_p$ \\
\midrule
MPCC     & $+12.0\%$ & $+71.5\%$ & $+121\%$ & $+17.5\%$ & $+26.2\%$ & $-5.2\%$ \\
\DQMPCC{}  & $\mathbf{+7.4\%}$ & $\mathbf{-10.1\%}$ & $\mathbf{+9.7\%}$ & $\mathbf{-4.1\%}$ & $\mathbf{+13.9\%}$ & $\mathbf{-4.8\%}$ \\
\bottomrule
\end{tabular}
\end{table}

Table~\ref{tab:sim2real} and Fig.~\ref{fig:pareto} show the \revr{transfer without retuning}: \DQMPCC{}'s sim-to-real changes in
gate-crossing offset, contour error, and lap time remain small, whereas
the baseline's worst-case gate offset
increases by $71.5\%$, crossing $R_{\rm safe}$.

\section{Conclusion}
\label{sec:conclusion}

We presented \DQMPCC{}, a racing MPCC with dual-quaternion prediction in the body frame and tangent-space contouring errors
\revr{whose principal empirical result is the no-retuning transfer from SiL to the completed flights}.
\revr{Without retuning, \DQMPCC{} keeps
crossings of the completed flights within the geometric crossing margin
and is faster, while the baseline crosses that margin at its worst gate on four of its six flights.}
\revr{Both formulations describe the same physics, and because the deployed controllers were tuned independently, the comparison does not isolate the effect of the representation.}
Limitations are motion-capture estimation, one confined circuit,
few flights, and no safety guarantee\revr{; component effects and estimator bias, latency, drift, and outliers
are not isolated}. Future work targets
onboard estimation, \revr{larger corridors, and an online attitude
reference consistent with the timing}.

\bibliographystyle{IEEEtran}

\bibliography{paper}

\vfill
\end{document}